\documentclass[11pt,a4paper]{article}
\usepackage[hyperref]{acl2019}
\usepackage{times}
\usepackage{latexsym}
\usepackage{graphicx}
\usepackage[labelformat=simple]{subcaption}

\usepackage{tabu}
\usepackage{multirow}
\usepackage{url}
\usepackage{amsmath}
\usepackage{xr}
\newcommand{\argmax}{\mathop{\mbox{argmax}}}

\newcommand{\equref}[1]{(\ref{#1})}

\aclfinalcopy 

\title{Content Word-based Sentence Decoding and Evaluating for Open-domain Neural Response Generation}

\author{Tianyu Zhao \\
\\
\\
\\\And
Shinsuke Mori \\
Graduate School of Informatics \\
Kyoto University \\
{\tt zhao@sap.ist.i.kyoto-u.ac.jp} \\\And
Tatsuya Kawahara \\
\\
\\
\\}

\date{}

\begin{document}
\maketitle

\begin{abstract}
Various encoder-decoder models have been applied to response generation in open-domain dialogs, but a majority of conventional models directly learn a mapping from lexical input to lexical output without explicitly modeling intermediate representations. Utilizing language hierarchy and modeling intermediate information have been shown to benefit many language understanding and generation tasks. Motivated by \textit{Broca's aphasia}, we propose to use a content word sequence as an intermediate representation for open-domain response generation. Experimental results show that the proposed method improves content relatedness of produced responses, and our models can often choose correct grammar for generated content words. Meanwhile, instead of evaluating complete sentences, we propose to compute conventional metrics on content word sequences, which is a better indicator of content relevance. 
\end{abstract}

\section{Introduction}
Hierarchical structure is an important feature in natural language, where low-level elements are combined to construct elements at a higher level. For example, a syntactic parse tree is built upon lexical tokens. Hierarchical structure is useful in various natural language processing (NLP) tasks. In natural language understanding (NLU), for example, low-level representations can be utilized to predict high-level representations~\citep{he2018syntaxsrl, qi2018universal, dong2018semantic}. In natural language generation (NLG) tasks, high-level representations can guide the decoding of produced text for better quality~\citep{dyer2016recurrent, kuncoro2018syntax}. 

We focus on response generation in text-based dialogs, where interlocutors exchange information via written language. \citet{pickering2004toward} showed that two symmetric processes of language comprehension and language production happen through multiple representations from lexical level (phonological level for spoken dialogs) to pragmatic level.

The conventional approach to neural response generation in open-domain dialogs, however, uses variants of encoder-decoder model to directly learn a mapping function from input text to output text~\citep{vinyals2015neural,serban2016building}, where high-level semantics are not explicitly modeled. Though modern neural networks are capable of generating fluent short sentences, the lack of high-level guidance prevents them from producing quality text with consistency and soundness. 

Different from open-domain dialogs, semantic representation is commonly used in task-oriented dialogs~\citep{wen2015semantically}. By using intentions (dialog act and slot-value pairs) as semantic constraints, generated responses are more rational and better controlled.~\citet{wen2017latent} showed that the success of task-oriented dialog models heavily relies on modeling the semantic representations. High-level representation such as intention is useful but usually unavailable in open-domain dialogs because annotation of intention requires domain-specific knowledge. Therefore, we hope to define a representation which can be easily obtained for open-domain dialogs. 

Motivated by \textit{Broca's aphasia}~\footnote{\url{https://en.wikipedia.org/wiki/Expressive_aphasia}}, we find that \textit{content word sequence} can be naturally introduced to response generation. \textit{Broca's aphasia} is 
\begin{quote}
``\textit{a type of aphasia characterized by partial loss of the ability to produce language caused by acquired damage to the anterior regions of the brain}''. 
\end{quote}
A patient with \textit{Broca's aphasia} usually includes only content words and omits function words that have little lexical meanings when producing a sentence. For instance, a patient may say ``\textit{walk dog}'' to express ``\textit{I will take the dog for a walk}''.

In this paper, we propose to identify two processes in response generation. Unlike the conventional approach, which directly predicts a response sentence $y$ given an input sentence $x$, we first predict a sequence of content words $c$ (e.g. ``\textit{walk dog}''), then decode a complete response sentence $y$ (e.g. ``\textit{I will take the dog for a walk}'') from the content words $c$. Our contribution is two-fold. (1) We propose to use a content word sequence as an intermediate representation for response generation in open-domain dialogs, which models the speech production process in \textit{Broca's aphasia}. (2) We propose to compute conventional evaluation metrics on content word sequences, which is a better indicator of content relevance.

The rest of this paper is organized as following. In Section~\ref{sec:related_works}, we discuss related works on neural response generation and applications of intermediate representations of language. We present the proposed models in Section~\ref{sec:method} and evaluation metrics in Section~\ref{sec:evaluation}. Experimental settings are given in Section~\ref{sec:conditions}. Results and analyses are given in Section~\ref{sec:experiment}. In Section~\ref{sec:conclusion}, we conclude with a brief summary and suggest future lines for further improvement of the proposed method.

\section{Related Works}
\label{sec:related_works}

Utilizing representations from other levels of the language hierarchy is a common practice in NLP researches, and it has been shown to benefit various tasks. In the recently published dependency parsing model of \textit{StanfordNLP}~\citep{qi2018universal}, representations of upstream tasks (segmentation, POS tagging, morphological tagging, and lemmatization) are used as inputs to dependency parsing. \citet{he2018syntaxsrl} confirmed the importance of syntactic representations in semantic role labeling (SRL) task. These works used multiple representations as input features but did not directly model their structures. \citet{dong2018semantic} proposed a coarse-to-fine process for semantic parsing. To generate a semantic representation, they use a sketch as an intermediate representation, which contains overall structure and glosses over low-level details such as variable names.

In generation tasks, high-level representations can guide the decoding of low-level surface text. \citet{ji2016latent} introduced a latent variable to discourse-level language model, which models inter-sentence relations in discourses or dialogs. And performances on both relation classification and language modeling tasks are significantly improved. \citet{dyer2016recurrent} proposed recurrent neural network grammars (RNNGs) to generate a sentence while simultaneously generating its corresponding parsing tree. Integrating syntax modeling with sentence generating helps the model outperform all previous language models. Following works~\citep{kuncoro2017recurrent, kuncoro2018syntax} further confirmed the contribution of explicit modeling composition and syntax to successful language models. 

Besides language modeling, high-level representations are also helpful in other tasks, especially where logical reasoning is considered important. In automated story generation, a model is supposed to complement a story given preceding context. \citet{martin2018event} decomposed the problem into generating a sequence of \textit{event}s and decoding natural language sentences conditioned on \textit{event}s, where an \textit{event} is a 4-tuple representation that contains a verb, a subject, an object, and extra information.

Back to the problem of neural response generation in dialogs, structured semantic representation is common in task-oriented dialog systems. An intent label accompanied with its slot-value pairs is used to describe the intention of a sentence (e.g. \textit{Inform(name=Seven\_days, food=Chinese)} means the sentence to be generated should inform its user of a Chinese restaurant named \textit{Seven days}). \citet{wen2015semantically} proposed using DA-gated Long Short-term Memory (LSTM) cell to make response generation conditioned on input semantics. Intentions can also be modeled as latent variables to allow training on unlabeled corpora~\citep{wen2017latent}.

The structured semantic representation used in task-oriented dialog systems can put strong constraints on generated responses, and thus is also helpful to control generation in open-domain dialogs. However, it requires (1) manual annotation of domain-specific DAs and slots, and (2) an external knowledge base to process user intentions and output system intention. And such resources are unavailable in open-domain systems. 

The most mentioned baseline models in open-domain response generation are vanilla encoder decoder~\citep{vinyals2015neural}, hierarchical encoder decoder~\citep{serban2016building}, and their counterparts with attentional mechanism. These models encode lexical inputs as a context vector, and decode lexical outputs from the vector. Merely learning from lexical representation results in many problems such as the lacks of diversity~\citep{li2016diversity}, persona consistency~\citep{li2016persona}, and discourse coherence~\citep{li2017neural}. Various methods have been proposed to control the decoding process by using extra representations as conditions. \citet{zhao2017learning} applied DA to control sentence type (e.g. \textsc{Yes-No Question} and \textsc{Statement-opinion}). DA is an abstractive representation that describes sentence function, so it can hardly affect the content of generated response. Instead keywords can identify current topic of a dialog, which is more content-related. \citet{yao2017towards} used pointwise mutual information (PMI) to extract keyword, and \citet{yu2018rich} used external tool along with a memory network to generate a keyword. The keywords are then used as auxiliary inputs to their response decoders. \citet{wu2018dynamic} proposed to control content by forcing decoder to use a predicted smaller vocabulary. Similar to our proposal, they also categorize words into function words and content words. To generate a response, they predict from context a limited number of content words. Lastly, \citet{serban2017multiresolution} proposed to generate two sequences by using multiresolution RNN (MrRNN). A coarse sequence that captures a compositional structure and semantics is first decoded, then a complete response sentence is decoded conditioned on the coarse sequence. They defined a coarse representation with nouns or activities-entities.

Our proposed content word-based sentence decoding shares common aspects with these past works, so we highlight the differences here. Both \citet{dong2018semantic} and \citet{serban2017multiresolution} used a coarse sequence as an intermediate representations and applied a two-step decoding process as we do. But their definitions of the coarse sequence vary with tasks. We define an intermediate representation as a sequence of content words that is to be included in the final sentence. This definition is universal, and models the process of human speech production as motivated by \textit{Broca's aphasia}, which provides us with solid theoretical support.


\section{Methodology}
\label{sec:method}

\begin{figure*}[!t]
\centering
\includegraphics[width=1.0\textwidth]{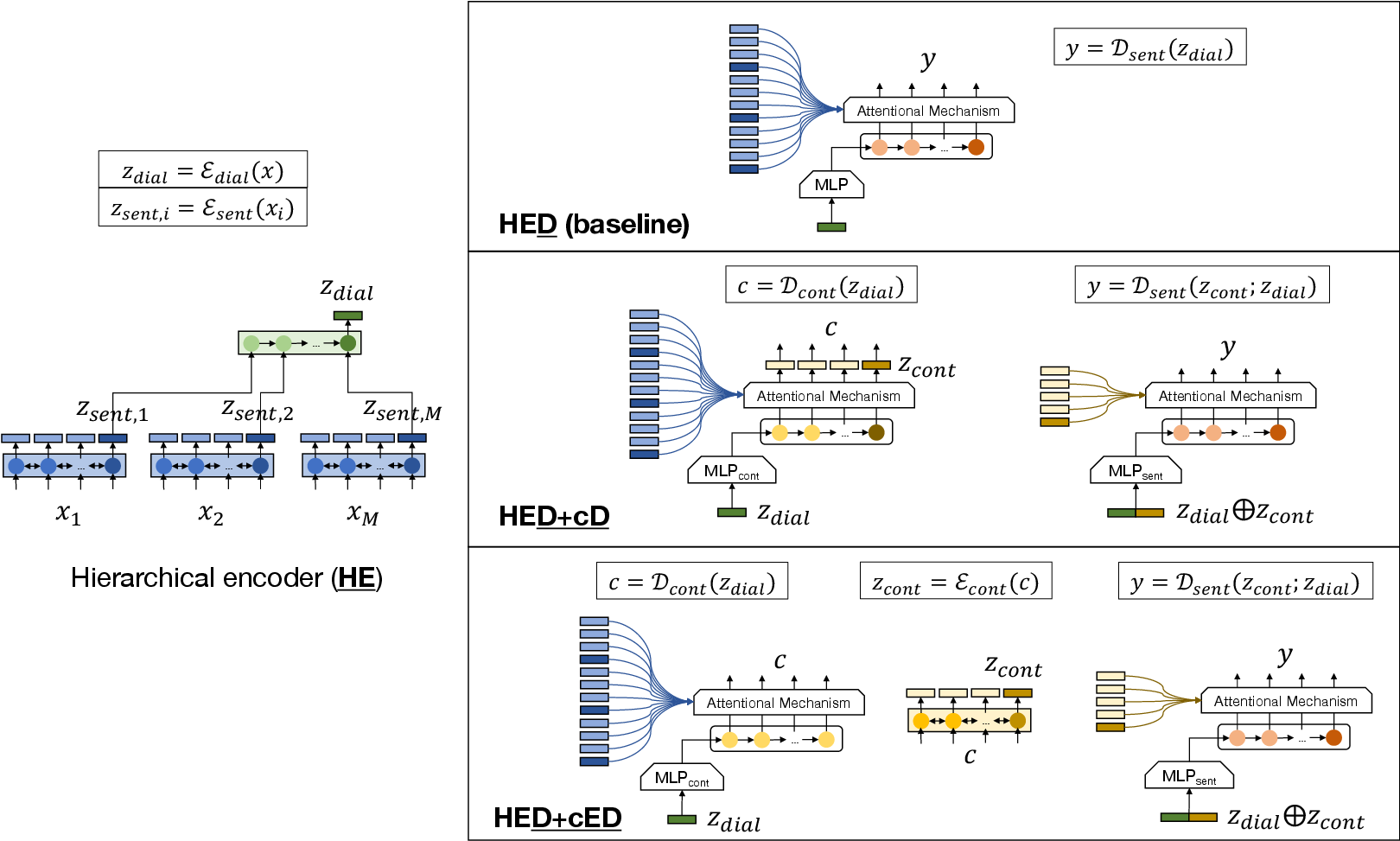}
\caption{The baseline model (HED) and the proposed models (HED+cD and HED+cED).}
\label{fig:models}
\end{figure*}

We first give an overall formulation of neural response generation task. Given input dialog context $x=(x_{1}, x_{2}, ..., x_{M})$, which is a list of preceding sentences from a dialog, where a sentence is a list of tokens $x_{i}=(w_{i,1}, w_{i,2}, ..., w_{i,N_{i}})$. A model produces a response sentence $y=(w_{1}, w_{2}, ..., w_{L})$. $M$, $N_{i}$, and $L$ are the lengths of the context, the $i$-th sentence in context, and the response, respectively. Since we use a left-to-right unidirectional RNN to decode the response, our basic objective function is:
\begin{align}
    P(y|x) = \prod_{n=1}^L P(w_{n}|w_{<n}, x).
\end{align}
Variations of the objective function are derived for different models in following subsections.

\subsection{Baseline: Hierarchical Encoder-Decoder}
\label{ssec:hed}
We choose \textbf{H}ierarchical \textbf{E}ncoder-\textbf{D}ecoder (HED)~\citep{serban2016building} as a representative of conventional models. As shown in Figure~\ref{fig:models}, HED uses sentence encoder $\mathcal{E}_{sent}$ to encode a sentence $x_{i}$ into a numeric vector $z_{sent, i}$. A dialog encoder $\mathcal{E}_{dial}$ encodes a entire dialog $x$ into $z_{dial}$.
\begin{align}
    \begin{split}
        z_{sent, i} 
        &= \mathcal{E}_{sent} (x_{i}) \\
    \end{split} \label{equ:sent_encode}\\
    \begin{split}
        z_{dial} 
        &= \mathcal{E}_{dial} (z_{sent, 1}, z_{sent, 2}, ..., z_{sent, M}).
    \end{split} \label{equ:dial_encode}
\end{align}

The encoders $\mathcal{E}_{sent}$ and $\mathcal{E}_{dial}$ are implemented as a bidirectional gated recurrent unit (BiGRU) network and a unidirectional GRU network~\citep{cho2014gru}, respectively. We use the last hidden state of $\mathcal{E}_{sent}$, which summarizes the input sentence, as sentence encoding $z_{sent, i}$. Similarly, dialog encoding vector $z_{dial}$ is the last hidden state of $\mathcal{E}_{dial}$.

A unidirectional GRU decoder $\mathcal{D}_{sent}$ generates response $y$ conditioned on dialog encoding $z_{dial}$. A small multilayer perceptron (MLP) network first transforms $z_{dial}$ into the initial hidden state of $\mathcal{D}_{sent}$. 
\begin{align}
    h_0^{{D}_{sent}} 
    = \text{MLP}( z_{dial} ).
\end{align}
Then we apply attentional mechanism to the decoder so that it can attend to hidden states of sentence encoders $h_{1:M,}^{\mathcal{E}_{sent}}$ during decoding.
\begin{align}
    y
    &= \argmax_{y} P (y | x), \\
    &= \argmax_{w_1, ..., w_N} \prod_{n=1}^L P(w_{n}|w_{<n}; x), \\
    &= \argmax_{w_1, ..., w_N} \prod_{n=1}^L \mathcal{D}_{sent} (w_{n}|w_{<n}; h_0^{{D}_{sent}}; h_{1:M,}^{\mathcal{E}_{sent}}).
\end{align}

\subsection{Content Word Sequence}
\label{ssec:content_words}

As mentioned before, behaviours of \textit{Broca's aphasia} patients suggest that we can apply a two-step decoding process for response generation. The first step produces a sequence of content words $c$, and the second fills in function words to form the final sentence $y$. The proposed model is trained to maximize objective function $P(y,c|x)$ on training data:
\begin{align}
    P(y,c|x) 
    &= P(y|c,x) \times P(c|x), \label{equ:ched} \\
    c
    &= (c_1, c_2, ..., c_{L_{cont}}),
\end{align}
where $c_i$ is the $i$-th content word and $L_{cont}$ is the length of the content word sequence.
 
Since most dialog corpora for response generation only contain $(x, y)$ pairs, we need to extract a content word sequence $c$ for each training sample and construct $(x, c, y)$ triplets. Here we explain the procedure of extracting the content word sequence. 

We first define function words because there is only a small number of them. Then we filter out function words in a sentence to obtain its corresponding content word sequence. Function words can be identified by their part-of-speech (POS) tags. Following categorization by Wikipedia,\footnote{\url{https://en.wikipedia.org/wiki/Function_word}} we use words that belong to \textsc{article}, \textsc{pronoun}, \textsc{preposition}, \textsc{conjunction}, \textsc{auxiliary verb}, \textsc{interjection}, or \textsc{particle}.  We also regard \textsc{punctuation}s as function words.

One alternative to locate function words in a sentence is to automatically associate tokens with POS tags using a tagger, and find function words with wanted POS types. But we found that the resulting sequences are noisy because of the imperfect performance of existing POS tagger tools. Therefore, we manually construct a vocabulary $V_{func}$ that contains words that commonly belong to the mentioned POS types. We exclude \textsc{pronoun}s and \textsc{punctuation}s from $V_{func}$ when constructing content word sequences during training because we notice that including \textsc{pronoun}s and \textsc{punctuation}s in content word sequences can significantly improve performances of the proposed models.

\begin{figure}
\centering
\includegraphics[width=1.0\columnwidth]{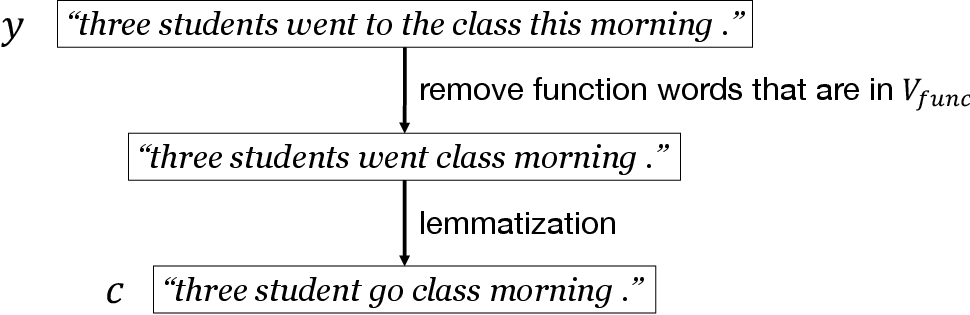}
\caption{Extract a content word sequence $c$ from a response sentence $y$.}
\label{fig:build_c}
\end{figure}

Given the function word vocabulary $V_{func}$ and a response sentence $w_1, ..., w_n$, we remove all function words $w_i \in V_{func}$ from the sentence. Instead of using left words directly, we lemmatize them using lemmatization tool by \textit{StanfordNLP}, and use their lemmas as content words. The reason behind it is that inflections such as ``\textit{-s}'' and ``\textit{-ing}'' usually serve a grammatical purpose, and we can remove such inflections and focus on the content-related stems. An example of the procedure is shown in Figure~\ref{fig:build_c}.

\subsection{Content Word-based Hierarchical Encoder-Decoder}
\label{ssec:ched}
To incorporate a content word sequence $c$ into the objective function as in Equation~\equref{equ:ched}. We need to add a component that generates $c$ and a mechanism that allows the decoding of $y$ to be conditioned on generated $c$. As shown in Figure~\ref{fig:models}, we extend the baseline HED to two variants: HED with \textbf{c}ontent \textbf{D}ecoder (HED+cD) and HED with \textbf{c}ontent \textbf{E}ncoder-\textbf{D}ecoder (HED+cED).

Both HED+cD and HED+cED use hierarchical encoders $\mathcal{E}_{sent}$ and $\mathcal{E}_{dial}$ to encode input context $x$ as in Equations~\equref{equ:sent_encode}-\equref{equ:dial_encode}, but they have different decoding processes.

\subsubsection{HED+cD}
Instead of decoding $y$ from $x$, HED-cD first decodes $c$ from $x$ using a content decoder $\mathcal{D}_{cont}$, which is also implemented as a unidirectional GRU network. We first use $\text{MLP}_{cont}$ to transform $z_{dial}$ to the initial hidden state of $\mathcal{D}_{cont}$.
\begin{align}
    & h_0^{{D}_{cont}} 
    = \text{MLP}_{cont}( z_{dial} ).
\end{align}
Then the decoding of $c$ is conditioned on its initial hidden state $h_0^{{D}_{cont}}$, and attends to hidden states of sentence encoders $h_{1:M,}^{\mathcal{E}_{sent}}$.
\begin{align}
    c
    &= \argmax_{c} P (c | x), \label{equ:decode_c_begin} \\
    &= \argmax_{c_1, ..., c_{L_{cont}}} \prod_{n=1}^{L_{cont}} P(c_{n}|c_{<n}; x), \\
    &= \argmax_{c_1, ..., c_{L_{cont}}} \prod_{n=1}^{L_{cont}} \mathcal{D}_{cont} (c_{n}|c_{<n}; h_0^{{D}_{cont}} ; h_{1:M,}^{\mathcal{E}_{sent}}). \label{equ:decode_c_end}
\end{align}

The second decoder $\mathcal{D}_{sent}$ generates $y$ from the outputs of encoders and the content decoder. Two kinds of information are needed to produce $y$, namely content-related and grammar-related information. To allow access to content information, we use attentional mechanism on $\mathcal{D}_{sent}$ to let it attend to hidden states of $\mathcal{D}_{cont}$, which are $h_{1:L_{cont}}^{\mathcal{D}_{cont}}$. Grammar information is needed (1) to decide which function words to use, and (2) to choose proper forms of content words (e.g. to choose between singular/plural forms). Thus, we extract grammar information from dialog encoding $z_{dial}$ and content encoding $z_{cont}$. We define $z_{cont}$ as the last hidden state of the content decoder. Another MLP network transform the concatenated vectors into the initial hidden state of $\mathcal{D}_{sent}$.
\begin{align}
    z_{cont} 
    &= h_{L_{cont}}^{\mathcal{D}_{cont}}, \\
    h_0^{{D}_{sent}} 
    &= \text{MLP}_{sent}( z_{dial} \oplus z_{cont}),
\end{align}
where $\oplus$ is concatenation operation. Then $\mathcal{D}_{sent}$ decodes $y$ as following:
\begin{align}
    y
    &= \argmax_{y} P (y | c,x), \\
    &= \argmax_{w_1, ..., w_N} \prod_{n=1}^{N} P(w_{n}|w_{<n}; c; x), \\
    &= \argmax_{w_1, ..., w_N} \prod_{n=1}^{N} \mathcal{D}_{sent} (w_{n}|w_{<n}; h_0^{{D}_{sent}} ; h_{1:L_{cont}}^{\mathcal{D}_{cont}}).
\end{align}

\subsubsection{HED+cED}
The second variation HED+cED extends HED+cD to have an extra content word sequence encoder $\mathcal{E}_{cont}$ for encoding generated $c$. 

Same as HED+cD, we obtain dialog encoding $z_{dial}$ following Equations~\equref{equ:sent_encode}-\equref{equ:dial_encode} and a content word sequence $c$ following Equations~\equref{equ:decode_c_begin}-\equref{equ:decode_c_end}. However, content encoding $z_{cont}$ and attention context of $\mathcal{D}_{sent}$ come from different sources.

A BiGRU-based content encoder $\mathcal{E}_{cont}$ takes $c$ as input. Its hidden states are then used to construct $z_{cont}$ and attention context for sentence decoding. Let $h_{1:L_{cont}}^{\mathcal{E}_{cont}}$ denotes the hidden states of $\mathcal{E}_{cont}$. We use the final hidden state as $z_{cont}$:
\begin{align}
    z_{cont} 
    &= h_{L_{cont}}^{\mathcal{E}_{cont}}, \\
    h_0^{{D}_{sent}} 
    &= \text{MLP}_{sent}( z_{dial} \oplus z_{cont}).
\end{align}
Then $y$ is decoded from the new initial hidden state and $\mathcal{D}_{sent}$ attends to $\mathcal{E}_{cont}$'s hidden states. 
\begin{align}
    y
    &= \argmax_{w_1, ..., w_N} \prod_{n=1}^{N} \mathcal{D}_{sent} (w_{n}|w_{<n}; h_0^{{D}_{sent}} ; h_{1:L_{cont}}^{\mathcal{E}_{cont}}).
\end{align}

\subsubsection{Inject Noise into Content Word Sequences}

In inference phase, content decoders can produce erroneous content word sequences, from which it is difficult to generate plausible responses. To improve the robustness of the proposed HED+cD and HED+cED, we inject noise into content word sequences during training, and force the models to recover complete responses from the noisy word content sequences. Therefore, we apply one of three operations to each content word sequence $c$: (1) Remove: randomly remove a word from $c$, (2) Repeat: repeat a random word in $c$ right after the word, (3) Insert: insert a random word at a random position into $c$. We choose an operation from the three randomly following a uniform probability distribution.

\section{Evaluation Metrics}
\label{sec:evaluation}

Conventional evaluation metrics in dialog response generation include sentence-level BLEU scores, word embedding similarities, and number of \textit{n}-gram types. We first consider the following measures:

\begin{itemize}
    \item \textbf{Sentence-level BLEU scores} - We use \textbf{B1} and \textbf{B2} to denote BLEU scores of unigram and bigram, respectively. BLEU scores of higher order \textit{n}-grams were too low to be informative.
    \item \textbf{Sentence-level word embedding similarities} - Cosine distance between a reference sentence and a hypothesis sentence in word embedding space. We use Embedding Average (\textbf{A-emb.}), Embedding Extrema (\textbf{E-emb.}), and Embedding Greedy (\textbf{G-emb.}) following previous works~\citep{serban2016building,zhao2017learning}. We use 200-dimensional \textit{Glove}~\citep{pennington2014glove} word embeddings pretrained on Twitter~\footnote{\url{https://nlp.stanford.edu/projects/glove/}} in evaluation.
    \item \textbf{Corpus-level distinct \textit{n}-gram counts} - To assess diversity of generated responses, we use \textbf{Dist-1} and \textbf{Dist-2} to denote numbers of distinct uni-grams and bi-grams at corpus level~\citep{li2016diversity}. 
\end{itemize}

\citet{liu2016not} showed that many of the previous metrics correlate poorly with human judgement. So instead of comparing a reference sentence $y$ with a hypothesis sentence $\hat{y}$, we propose to compare their corresponding content word sequences $c$ and $\hat{c}$. The content word sequence versions of the previous measures are prefixed by \textbf{c-}. Therefore, we will report content word sequence measures \textbf{cB1}, \textbf{cB2}, \textbf{cA-emb.}, \textbf{cE-emb.}, \textbf{cG-emb.}, \textbf{cDist-1}, and \textbf{cDist-2}, while the original measures are also reported as a reference. Additionally, we calculate the coverage ratio of content word types \textbf{cCoverage}:
\begin{align*}
    \mbox{cCoverage}(c,\hat{c}) = 
    \frac{|\{t | t \in c \land t \in \hat{c}\}|}
    {|\{t | t \in c\}|},
\end{align*}
where $t$ is a word type.

We use an example to show that the proposed metrics reflect content relevance better than the original metrics in Table~\ref{tab:eval_example}, where we compare the scores of two hypotheses given a reference sentence. Sentence texts can be found in the table's caption. The results show that hypothesis 1 is better according to the original metrics, while hypothesis 2 is more similar to the reference according to the proposed metrics. The example suggests that the new metrics correlate better with our judgement in general.

\begin{table}[!t]
    \centering
    \begin{tabu} to 1.0\columnwidth {c|cc}
        \textbf{Metric \%} & \textbf{hypothesis 1} & \textbf{hypothesis 2} \\
        \hline
        \hline
        B1 & \textbf{23.08} & 7.59 \\
        \hline
        B2 & \textbf{8.33} & 0.0 \\
        \hline
        E-Emb. & 46.90 & \textbf{57.21} \\
        \hline
        A-Emb. & \textbf{94.70} & 92.66 \\
        \hline
        G-Emb. & \textbf{78.51} & 75.67 \\
        \hline
        \hline
        cB1 & 25.00 & \textbf{47.77} \\
        \hline
        cB2 & 0.0 & 0.0 \\
        \hline
        cE-Emb. & 44.83 & \textbf{73.79} \\
        \hline
        cA-Emb. & 80.12 & \textbf{86.27} \\
        \hline
        cG-Emb. & 67.12 & \textbf{78.88} \\
        \hline
        cCoverage & 25.00 & \textbf{50.00}
    \end{tabu}
    \caption{An example to explain the difference between evaluating complete sentences and evaluating content word sequences. The reference is ``\textit{do you have any skirt that will go with this sweater ?}'', hypothesis 1 is ``\textit{he will leave tomorrow but he does not have any plan yet .}'', and hypothesis 2 is ``\textit{the skirts match well with these sweaters .}''.}
    \label{tab:eval_example}
\end{table}

\section{Experimental Conditions}
\label{sec:conditions}
For training purpose we used two corpora, namely DailyDialog~\citep{li2017dailydialog} and the Cornell Movie Dialog Corpus~\citep{danescu2011chameleons} (we will refer to it as CornellMovie for brevity in the following text). Both datasets assume speaker switch between two adjacent turns, so there are some very long turns~\footnote{We found in DailyDialog, some turns contain more than 100 tokens, and in CornellMovie, a turn contains more than 1000 tokens.}. Thus, we conducted sentence segmentation in preprocessing to split a long turn into sentences with smaller lengths, and sentences with more than 40 tokens were truncated. We used a recently published NLP tool \textit{StanfordNLP}~\footnote{{\url{https://stanfordnlp.github.io/stanfordnlp/}}} to apply a series of preprocessing to the datasets: sentence segmentation, tokenization, and lemmatization of content words. We defined the 10,000 most frequent words as the vocabulary for DailyDialog and 20,000 for CornellMovie. Then we splitted the datasets into training/development/test sets at a ratio of 0.8/0.1/0.1. The resulting preprocessed DailyDialog dataset has 104k/17k/14k sentences for training/development/test, and CornellMovie dataset has 474k/26k/27k sentences. The average sentence lengths are 11.00 and 9.91 in the two datasets, respectively. We give implementation details in Appendix \ref{sec:implementation} and the constructed function word vocabulary in Appendix \ref{sec:app_func_word} due to the limit of pages.


\section{Results and Analyses}
\label{sec:experiment}

\subsection{Quantitative Analysis}

To assess the benefit of introducing the content word sequence representation, we compare the proposed model HED+cD and HED+cED with a baseline HED. A vanilla \textbf{HED w/o attn}, which is the non-attentional version of HED is also added for reference. 

\begin{table}[!t]
    \centering
    \begin{tabu} to 1.0\columnwidth {c|cccc}
        \multirow{2}{*}{\textbf{Metric \%}} & \textbf{HED} & \multirow{2}{*}{\textbf{HED}} & \textbf{HED} & \textbf{HED} \\
        & \textbf{w/o attn} && \textbf{+cD} & \textbf{+cED} \\
        \hline
        \hline
        B1 & 12.86 & 13.56 & \textbf{14.07} & 13.69 \\
        \hline
        B2 & 1.46 & \textbf{1.67} & 1.56 & 1.54 \\
        \hline
        E-Emb. & 54.02 & 54.36 & \textbf{55.29} & 54.81 \\
        \hline
        A-Emb. & 89.69 & 89.95 & \textbf{90.11} & 89.84 \\
        \hline
        G-Emb. & 75.57 & 75.75 & \textbf{76.16} & 75.95 \\
        \hline
        Dist-1 & 2570 & \textbf{3098} & 2660 & 2901 \\
        \hline
        Dist-2 & 13597 & 19056 & 17110 & \textbf{19357} \\
        \hline
        \hline
        cB1 & 2.01 & 2.53 & \textbf{3.13} & 2.97 \\
        \hline
        cB2 & 0.14 & 0.18 & 0.28 & \textbf{0.30} \\
        \hline
        cE-Emb. & 44.25 & 44.89 & \textbf{47.81} & 46.43 \\
        \hline
        cA-Emb. & 61.76 & 62.74 & \textbf{65.01} & 64.21 \\
        \hline
        cG-Emb. & 53.04 & 53.88 & \textbf{56.50} & 55.34 \\
        \hline
        cDist-1 & 2016 & \textbf{2385} & 1981 & 2189 \\
        \hline
        cDist-2 & 8490 & \textbf{11798} & 8749 & 11157 \\
        \hline
        cCoverage & 2.81 & 3.54 & \textbf{4.36} & 4.13 
    \end{tabu}
    \caption{Results on DailyDialog corpus. }
    \label{tab:dd_result}
\end{table}

\begin{table}[!t]
    \centering
    \begin{tabu} to 1.0\columnwidth {c|cccc}
        \multirow{2}{*}{\textbf{Metric \%}} & \textbf{HED} & \multirow{2}{*}{\textbf{HED}} & \textbf{HED} & \textbf{HED} \\
        & \textbf{w/o attn} && \textbf{+cD} & \textbf{+cED} \\
        \hline
        \hline
        B1 & 11.37 & \textbf{12.04} & 11.97 & 11.47 \\
        \hline
        B2 & 1.15 & \textbf{1.51} & 0.93 & 0.99 \\
        \hline
        E-Emb. & 52.07 & 52.45 & \textbf{53.37} & 53.27 \\
        \hline
        A-Emb. & 84.24 & 84.53 & \textbf{85.64} & 84.13 \\
        \hline
        G-Emb. & 70.24 & 70.59 & \textbf{71.44} & 70.82 \\
        \hline
        Dist-1 & 2414 & \textbf{3370} & 2638 & 2846 \\
        \hline
        Dist-2 & 10110 & \textbf{16279} & 14044 & 14411 \\
        \hline
        \hline
        cB1 & 1.76 & 2.43 & \textbf{2.82} & 2.74 \\
        \hline
        cB2 & 0.14 & 0.25 & 0.23 & \textbf{0.29} \\
        \hline
        cE-Emb. & 38.99 & 39.70 & \textbf{47.10} & 43.04 \\
        \hline
        cA-Emb. & 55.06 & 55.21 & \textbf{62.03} & 56.55 \\
        \hline
        cG-Emb. & 48.28 & 48.45 & \textbf{55.47} & 50.46 \\
        \hline
        cDist-1 & 1878 & \textbf{2612} & 1981 & 2178 \\
        \hline
        cDist-2 & 5232 & \textbf{8429} & 6005 & 6081 \\
        \hline
        cCoverage & 2.44 & 3.40 & \textbf{4.04} & 3.97
    \end{tabu}
    \caption{Results on CornellMovie corpus.}
    \label{tab:cm_result}
\end{table}

Table~\ref{tab:dd_result} gives the results on DailyDialog corpus. In the first group of original metrics, HED+cD reaches the best B1, E-emb., A-emb., and G-emb. scores. In the second group of metrics, HED+cD and HED+cED have even larger improvements over the baselines. HED+cD achieves relative improvements of $23.72\%$ in cB1., $6.50\%$ in cE-Emb., $3.62\%$ in cA-Emb., $4.86\%$ in cG-Emb., and $23.16\%$ in cCoverage in comparison with HED, which suggests that explicitly modeling content word sequence can improve content relevance. The HED baseline reaches a higher B2 than others, but we have shown in Section~\ref{sec:evaluation} that B2 correlates poorly with human judgement. HED also has higher diversity scores, but HED+cED is only slightly worse than HED. 

Table~\ref{tab:cm_result} shows consistent results on CornellMovie corpus. The proposed models even outperform the baselines with larger margins in embedding-based metrics. HED+cD achieves relative improvements of $16.05\%$ in cB1., $18.39\%$ in cE-Emb., $12.35\%$ in cA-Emb., $14.49\%$ in cG-Emb., and $18.82\%$ in cCoverage in comparison with HED. HED+cED is not as good as HED+cD in most measurements, but it produces more diverse responses than HED+cD while achieving higher content relevance than the HED baseline.

\begin{table}
    \centering
    \tabulinesep=1.2mm
    \begin{tabu} to 0.9\columnwidth {c|c|X[c]}
        \multicolumn{2}{c|}{\textbf{Example}} & \textbf{Text} \\
        \hline
        \hline
        \multirow{2}{*}{\#1} & $\mathbf{\hat{c}}$ & \textit{you make any difference ?} \\
        \cline{2-3}
        & $\mathbf{\hat{y}}$ & \textit{have you made any difference ?} \\
        \hline
        \multirow{2}{*}{\#2} & $\mathbf{\hat{c}}$ & \textit{there also some good idea .} \\
        \cline{2-3}
        & $\mathbf{\hat{y}}$ & \textit{there is also some good ideas .} \\
    \end{tabu}
    \caption{Examples of grammar analysis.}
    \label{tab:grammar_check}
\end{table}

\begin{table}
    \centering
    \tabulinesep=1.2mm
    \begin{tabu} to 1.0\columnwidth {c|c|X[c,m]}
        \multicolumn{2}{c|}{\textbf{Example}} & \textbf{Text} \\
        \hline
        \hline
        \multirow{2}{*}{\#1} & predicted DA & \textsc{directive} \\
        \cline{2-3}
        & $\mathbf{\hat{y}}$ & \textit{you can meet us there tomorrow evening .} \\
        \hline
        \multirow{2}{*}{\#2} & predicted DA & \textsc{question} \\
        \cline{2-3}
        & $\mathbf{\hat{y}}$ & \textit{how long will it take ?} \\
    \end{tabu}
    \caption{Examples for DA type agreement.}
    \label{tab:da_check}
\end{table}

\subsection{Qualitative Analysis}

To confirm that the proposed method is able to model the two-step process of sentence production, we take HED+cED trained on DailyDialog as an example and analyze the behaviour of $\mathcal{D}_{sent}$. See more examples in Tables~\ref{tab:dd_examples} and \ref{tab:cm_examples} in Appendix.

\subsubsection{Correctness of Grammar}
To see whether $\mathcal{D}_{sent}$ can correctly fill in function words and choose proper forms for content words. We compare several generated responses $\hat{y}$ with generated content words $\hat{c}$ on which they are conditioned. In example 1 from Table~\ref{tab:grammar_check}, $\mathcal{D}_{sent}$ is able to add a function word ``\textit{have}'' and replace ``\textit{make}'' with ``\textit{made}'' to form a grammatically correct sentence. However, there are also some cases like in example 2, where the form of the transformed word ``\textit{ideas}'' does not agree with function word ``\textit{is}''.

\subsubsection{Correctness of Dialog Act}

To check whether grammatical information is encoded in the initial hidden state $h_0^{{D}_{sent}}$, we add an extra MLP network to predict the response's DA label from $h_0^{{D}_{sent}}$. An extra cross entropy loss of DA classification is also added to the objective function. While sampling a response $\hat{y}$, we also record the DA label predicted from $h_0^{{D}_{sent}}$ and examine whether $\hat{y}$ conforms to the DA type~\footnote{DailyDialog is annotated with four DA types, namely \textsc{inform}, \textsc{question}, \textsc{directive}, and \textsc{commissive}.}. Examples in Table~\ref{tab:da_check} suggest that $\hat{y}$ does reflect the DA types encoded in $h_0^{{D}_{sent}}$.

\section{Conclusion and Future Work}
\label{sec:conclusion}
Motivated by \textit{Broca's aphasia}, we proposed to use a content word sequence as an intermediate representation for open-domain response generation in this paper. We also proposed a new set of content word-based evaluation metrics, which better assesses the system's performance of content relevance. Experiments on two corpora showed that our method outperforms baselines in quantitative analysis, and it can construct grammatically correct sentences from content words.

The proposed models still have room for further improvements. Here we suggest possible ways of improvements in future works. (1) The content word sequence is a relatively low-level representation in speech production, and operations such as logical reasoning are almost impossible by merely using plain text and content word sequences. Structured semantics such as \textit{event} annotation~\citep{martin2018event} is promising for better dialog modeling. (2) Sentence decoders in HED+cD and HED+cED make grammatical errors sometimes. It can be improved by disentangling content and grammar information and providing explicit learning signals for using correct grammars, as \citet{hu2017toward} forced a latent variable to encode text attributes such as sentiment and tense via adversarial training.

\bibliography{../collection}
\bibliographystyle{acl_natbib}

\newpage
\appendix
\section{Implementation Details}
\label{sec:implementation}
We implement all models in \emph{PyTorch} and use the following hyperparameters. Word embeddings have 200 dimensions and are initialized by \emph{GloVe} word embeddings pretrained on Twitter. Within a model, word embeddings are shared by encoders and decoders. Weight tying is also applied to share parameters between word embeddings and decoders' output layers~\citep{press2017using}. The BiGRU encoders have a hidden size of 300, and the GRU decoders have a hidden size of 200. All GRU networks only have one hidden layer. Function word vocabularies are constructed following Section~\ref{ssec:content_words}, and the resulting vocabulary sizes are 195 for DailyDialog and 227 for CornellMovie. Window size of preceding sentences in context is 5. Weights other than word embeddings are initialized with values drawn from a uniform distribution $[-0.08, 0.08]$. All trainable parameters are optimized using Adam method~\citep{kingma2014adam} for 20 epochs with an initial learning rate of 0.0003. Batch sizes are 32 for DailyDialog and 64 for CornellMovie.

\begin{table*}[!t]
    \centering
    \tabulinesep=1.2mm
    \begin{tabu} to 1.0\textwidth {c|X[c,m]}
        \hline
        \textbf{Category} & \textbf{Function Words} \\
        \hline
        \hline
        \textsc{article} & a an the \\
        \hline
        \textsc{pronouns} & i me my mine myself you you your yours yourself he him his himself she her hers herself it it its itself we us our ours ourselves you you your yours yourselves they them their theirs themselves this these that those former latter that who whom which when where something anything nothing somewhere anywhere nowhere someone anyone none who what which whom whose where when why how \\
        \hline 
        \textsc{preposition} & about above across after against along among around as at before behind below beneath beside between beyond by despite down during except for from in inside into like near of off on onto opposite out outside over past round since than though to towards under underneath unlike until up upon via with within without \\
        \hline
        \textsc{conjunction} & for but yet both and either or neither nor whether after although as because before if lest once only since so supposing that than though till unless until when where wherever while whence whenever whereas whereby whereupon \\
        \hline
        \textsc{auxiliary} & be am are is was were being been can could dare do does did have has had having may might must need ought shall should will would \\
        \hline
        \textsc{interjection} & ah aha ahem alas amen aw aww bada bing bah bingo boo boo-hoo booyah bravo brr brrr bye bye-bye c'mon dang duh eh fiddledeedee gee fiddledeedee golly goodbye gosh ha hallelujah heigh-ho hello hey hi hiya hooray hmm hrm ho howdy huh ick jeez kaboom kapow mama mia nah nope oh ok okay ouch ow shh well woo-hoo wow yabba dabba doo yippee yummy \\
        \hline
        \textsc{particle} & yes no not n't 's yeah then \\
        \hline        
        \textsc{punctuation} & , . ! ? - ... ' \\
        \hline
    \end{tabu}
    \caption{Manually constructed function vocabulary.}
    \label{tab:func_vocabs}
\end{table*}

\section{Function word vocabulary}
\label{sec:app_func_word}

We use a manually constructed function word vocabulary as described in Section \ref{ssec:content_words}. The resulting vocabulary has some overlaps between different categories, and may have missed some words, but it generally works well in our experiments. Table~\ref{tab:func_vocabs} presents words in the function word vocabulary. The vocabulary is adapted to a corpus by removing words that do not occur in the corpus. Notice that \textsc{pronoun}s are excluded from the function word vocabulary when constructing content word sequences during training, and are included in the function word vocabulary in evaluation.

\renewcommand{\arraystretch}{1.5} 
\begin{table*}[!t]
    \centering
    \begin{tabular}{m{6.5cm}|m{2.5cm}|m{6cm}}
        \hline
        \textbf{Context} & \textbf{Reference} & \textbf{Response} \\
        \hline 
        \hline 
        \parbox[t]{6.5cm}{
            B: i 'm not sure , is this our stop ? \\
            A: yeah , get off the bus . \\
            B: do you have any idea where we are ? \\
            A: i 'm really not sure .
        }
        & 
        B: this ca n't be the right stop .
        &
        \parbox[t]{6cm}{
            \textbf{HED w/o attn:} what about the bus ? \\
            \textbf{HED:} i see . \\
            \textbf{HED+cD:} [you stop too fast .] \\
            you 've stopped too fast . \\
            \textbf{HED+cED:} [i think we stop bus .] \\
            i think we should stop by bus .
        }
        \\
        \hline
        \parbox[t]{6.5cm}{
            B: can you tell me the price for a bus pass ? \\
            A: it costs \$ 65 to buy a monthly pass . \\
            B: is there a cheaper bus pass ? \\
            A: there are student passes .
        }
        &
        B: for how much ?
        &
        \parbox[t]{6cm}{
            \textbf{HED w/o attn:} well , we have to take it and take care of the shuttle to the airport and take a shuttle to the airport at the airport . \\
            \textbf{HED:} can you tell me the fact that the price is satisfactory ? \\
            \textbf{HED+cD:} [how much it ?] \\
            how much is it ? \\
            \textbf{HED+cED:} [how much i pay ?] \\
            how much do i pay for ?
        }
        \\
        \hline
        \parbox[t]{6.5cm}{
            B: what do you need ? \\
            A: i need to use the internet . \\
            B: you have your library card , right ?
        }
        &
        A: yes , i do .
        &
        \parbox[t]{6cm}{
            \textbf{HED w/o attn:} it 's on the left . \\
            \textbf{HED:} yes , i have a card . \\
            \textbf{HED+cD:} [here you .] \\
            yes , here you are . \\
            \textbf{HED+cED:} [, i .] \\
            yes , i do .
        }
        \\
        \hline
        \parbox[t]{6.5cm}{
            B: i 'm deeply grateful for all you 've done . \\
            A: you 're welcome . 
        }
        &
        A: i enjoyed helping you .
        &
        \parbox[t]{6cm}{
            \textbf{HED w/o attn:} if i 'm in a hurry , i would n't be able to help you recover soon . \\
            \textbf{HED:} i hope you can make it . \\
            \textbf{HED+cD:} [i mention you any help .] \\
            i 'll mention you any help . \\
            \textbf{HED+cED:} [i appreciate you help .] \\
            i appreciate your help .
        }
        \\
        \hline
        \parbox[t]{6.5cm}{
            B: let 's meet before next week 's meeting , jane . \\
            B: i want to kick around some ideas with you . \\
            A: no problem .
        }
        &
        A: i 've got a really light week .
        &
        \parbox[t]{6cm}{
            \textbf{HED w/o attn:} is it all fun ? \\
            \textbf{HED:} what ? \\
            \textbf{HED+cD:} [i make plan meeting you .]\\
            i 'll make plans for you . \\
            \textbf{HED+cED:} [i see you tomorrow .]\\
            i 'll see you tomorrow .
        }
        \\
        \hline
        \parbox[t]{6.5cm}{
            A: i would like to do that . \\
            B: how much ? \\
            A: i only want to transfer \$ 100 . \\
            B: it 'll only take me a minute to complete your transaction . \\
            A: great .
        }
        &
        A: thank you very much .
        &
        \parbox[t]{6cm}{
            \textbf{HED w/o attn:} i 'll buy a new card . \\
            \textbf{HED:} i 'll take it . \\
            \textbf{HED+cD:} [i make agreement now .] \\
            i have made an agreement now . \\
            \textbf{HED+cED:} [i see you receipt ?] \\
            may i see your receipt ? 
        }
        \\
        \hline
    \end{tabular}
    \caption{Generated examples from DailyDialog. Texts in square brackets refer to content word sequences.}
    \label{tab:dd_examples}
\end{table*}

\begin{table*}[!t]
    \centering
    \begin{tabular}{m{6cm}|m{2.5cm}|m{7cm}}
        \hline
        \textbf{Context} & \textbf{Reference} & \textbf{Response} \\
        \hline
        \hline  
        \parbox[t]{6cm}{
            A: what are we for then ? \newline
            A: we 're UNK . \newline
            A: okay , buy a lot of stuff , you 're a good citizen . \newline
            A: but if you do n't buy a lot of stuff , you know what ? \newline
            A: you 're mentally ill !
        }
        & 
        A: that 's a fact !
        &
        \parbox[t]{7cm}{
            \textbf{HED w/o attn:} i 'm not going to be a hero . \\
            \textbf{HED:} i 'm not going to take you to the bathroom . \\
            \textbf{HED+cD:} [you get you own personal life .] \\
            you 've got your own personal life . \\
            \textbf{HED+cED:} [i .] \\
            i do n't .
        }
        \\
        \hline   
        \parbox[t]{6cm}{
            A: how do you know it 's a she ? \newline
            B: i saw her in the water . \newline
            B: beside me . \newline
            B: she was filling the bathtub . \newline
            A: what does she look like ?
        }
        & 
        B: pretty .
        &
        \parbox[t]{7cm}{
            \textbf{HED w/o attn:} i do n't know . \\
            \textbf{HED:} she 's a real crook . \\
            \textbf{HED+cD:} [she look .] \\
            she looked like a brunette . \\
            \textbf{HED+cED:} [she bathtub .] \\
            she 's in the bathtub .
        }
        \\
        \hline  
        \parbox[t]{6cm}{
            A: marry me . \\
            B: i thought you were almost engaged , stu ... ? \\
            A: you really love this guy ... ? \\
            B: he 's got great hands . \\
            A: hey - i love gloves - i look really great in gloves .
        }
        & 
        B: i like him .
        &
        \parbox[t]{7cm}{
            \textbf{HED w/o attn:} i 'm not going to be a reporter . \\
            \textbf{HED:} i 'm really sorry about that . \\
            \textbf{HED+cD:} [i know .] \\
            i do n't know about that , either . \\
            \textbf{HED+cED:}  [you dating you lillian , you know , you really love she ?] \\
            you 're not dating your lillian , you know , do you really love her ?
        }
        \\
        \hline  
        \parbox[t]{6cm}{
            B: how long have you been watching me ?
        }
        & 
        A: i have something i have to show you ...
        &
        \parbox[t]{7cm}{
            \textbf{HED w/o attn:} i do n't know . \\
            \textbf{HED:} i do n't know . \\
            \textbf{HED+cD:} [year .] \\
            two years . \\
            \textbf{HED+cED:}  [two hour .] \\
            two hours .
        }
        \\
        \hline  
        \parbox[t]{6cm}{
            A: the bullet is moving through the UNK UNK , killing off his sense of touch . \newline
            A: he feels no pain . 
        }
        & 
        A: he can push himself harder , longer than any normal man .
        &
        \parbox[t]{7cm}{
            \textbf{HED w/o attn:} he 's got a UNK . \\
            \textbf{HED:} the bullet is a UNK . \\
            \textbf{HED+cD:} [i think he kill he .] \\
            i think he kills him . \\
            \textbf{HED+cED:} [he get UNK .] \\
            he 's got UNK .
        }
        \\
        \hline  
        \parbox[t]{6cm}{
            A: you think i ca n't do it . \\
            A: you think i 'm going to screw up . \\
            A: or get myself killed . \\
            B: i think running this pipeline would be a difficult job for anyone . \\
            B: especially king 's daughter .
        }
        & 
        A: you are wrong , mr. bond .
        &
        \parbox[t]{7cm}{
            \textbf{HED w/o attn:}  i 'm going to tell her about it . \\
            \textbf{HED:} i 'm going to have to tell you about the policy . \\
            \textbf{HED+cD:} [i think i know what you .] \\
            i do n't think i know what you 're thinking . \\
            \textbf{HED+cED:} [he take advantage we .] \\
            he 's taken advantage of us .
        }
        \\
        \hline
    \end{tabular}
    \caption{Generated examples from CornellMovie. Texts in square brackets refer to content word sequences.}
    \label{tab:cm_examples}
\end{table*}

\end{document}